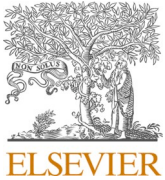
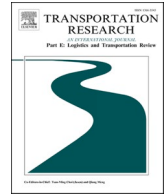

# Methodology for generating synthetic labeled datasets for visual container inspection

Guillem Delgado [*], Andoni Cortés, Sara García, Estíbaliz Loyo, Maialen Berasategi, Nerea Aranjuelo

*Vicomtech Foundation, Basque Research and Technology Alliance (BRTA), Mikeletegi 57, P. Tecnológico, 20009 San Sebastián, Spain*

A B S T R A C T

Nowadays, containerized freight transport is one of the most important transportation systems that is undergoing an automation process due to the Deep Learning success. However, it suffers from a lack of annotated data in order to incorporate state-of-the-art neural network models to its systems. In this paper we present an innovative methodology to generate a realistic, varied, balanced, and labelled dataset for visual inspection task of containers in a dock environment. In addition, we validate this methodology with multiple visual tasks recurrently found in the state of the art. We prove that the generated synthetic labelled dataset allows to train a deep neural network that can be used in a real world scenario. On the other side, using this methodology we provide the first open synthetic labelled dataset called SeaFront available in: https://datasets.vicomtech.org/di21-seafront/readme.txt.

## 1. Introduction

Containerized freight transport is one of the predominant cargoes transportation methods today due to its high capability of storage and organization. This field is not a stranger to automation as the different processes involved at each point of this transportation chain are undergoing a trend towards automation. The main task undergoing this change is the inspection of containers, more specifically the detection of the International Maritime Dangerous Goods (IMDG) markers, container marker identifier (ID markers) recognition and seal presence detection, among others. These are usually manual tasks done by the port operator during the loading and unloading of the containers with a Ship To Shore (STS) crane. In addition, visual check of containers being discharged from a vessel is performed, giving emphasis on possible structural damages in order to prevent disputes between transportation enterprises and wharf. Currently, it takes about 15 s for a person to enter all these data into a terminal operating system (TOS). In parallel, it takes up to 30 s for another worker to do the container inspection for possible damages and the seal presence check. The two activities mostly run in parallel, which means that on average the whole visual inspection takes about 30 s per container. Delays attributed to manual inspection of containers are directly related to unnecessary waiting times (STS crane idling, unevenly distributed movements on the yard, etc.). That means multiple people working in different tasks. With the raise and success of Deep Learning techniques in visual tasks, there has been some research tackling these issues, such as detection and segmentation of damages, IMDG marker detection, corrosion, and damages inspection and so on. However, due to the fact that public data from harbours is scarce and standardized data is even scarcer, there is a lot of time and cost that it is needed to devote in the data collection process. This process is even more difficult when taking into consideration all the possibilities during training as, in the vast majority of the cases, there will not be enough images to solve a particular problem in a specific scenario with a specific camera setup. In addition, due to the lack of effective tools and the annotation complexity, the dataset collection becomes even more difficult.

In this work we present the first methodology to build synthetic labelled data for training visual tasks in multi-camera visual inspection systems of containers. The proposed methodology focuses on containers in an STS crane. However, the system could be easily reused in different port context with little effort. We provide reproducible guidelines to generate synthetic data that can be really diverse using the proposed algorithms and can tackle multiple tasks such as detection and segmentation of container damages, detection and recognition of ID markers, detection, and classification of IMDG markers and classification of the images as door/no door. The generation of all the images and its labels are completely automatic and without any need of human intervention, as all the images' labels are generated with a precise ground truth automatically using the API of open-source tools and the algorithms proposed. Using this methodology, we provide the first open synthetic labelled dataset containing around 10,000 images of containers fully annotated for multiple tasks. Moreover, we provide a baseline for most of these tasks comparing it with state-of-the art models and a demonstration about how training with synthetic data affects the domain gap between real and synthetic data, comparing it to an existing setup in Luka Koper port, which is related to the European project that supports this research.

The rest of the paper is organized as follows; Section 2 describes the state of the art of detection and segmentation of containers, its damages and prior works related to synthetic data generation; Section 3 describes the methodology followed in the generation of the synthetic labelled dataset, statistics from the data and the labeling format; Section 4 introduces the different visual tasks used for the validation of the generated synthetic data, thus, proposing a baseline; Section 5 explains the experimentation and the results obtained; Section 6 shows how domain gap affects the models trained with synthetic data; finally, Section 7 present the conclusion and future lines of work.

## 2. Related work

There have been multiple research projects based on container's inspection, mostly focused on the detection of container's damages, segmentation and ID marker recognition. The first studies regarding damages detection using traditional computer vision appeared in 1995 with Nakazawa et al. (Nakazawa et al., 1163), where they proposed a method using illumination differences with traditional segmentation algorithms. Son et al. (Son et al.) proposed a new function called Capsize-Gaussian-Function in order to detect edges of damage or deformation in containers. In addition, identifier recognition has also been tackled with Kim et al. (Kim et al., 2006) where they proposed an automatic container identification system based on ART2 (Carpenter and Grossberg, 1987) architecture. Later studies developed systems focused on defining features for containers based on edge detections and pixel information, that checks automatically if the containers have been opened or exchanged (Huang et al., 2013).

Nowadays, Deep Learning has been used more and more in all the computer vision tasks and this is also true for the visual inspection of shipping containers within port contexts. Neural networks have proven to provide more accurate results compared to traditional methods. Verma et al. (Verma et al., 2016) tackled the issue of text recognition proposing a solution in the form of an end-to-end pipeline that uses Region Proposals for text detection in conjunction with Spatial Transformer Networks for text recognition. In addition, other existing networks have been proposed, such as Faster-RCNN (Ren et al., 2017), with a binary search tree to find the container identifier (Zhiming et al., 2019). A variation of Mask-RCNN has also been used in damage detection, giving place to Fmask-RCNN (Li et al., 2020) which introduces changes in the backbone, fusions in the FPN and multiple fully connected layers. Bahrami et al. (Bahrami et al., 2020) also introduced neural networks architectures to detect corrosion in containers. They used different models like Faster R-CNN, SSD-MobileNet and SSD Inception V2 and added an anchor box optimizer in order to detect and localize the corrosion. Classification is also present in damage classification, using a lightweight CNN (MobileNetv2) based on transfer learning, Wang et al. (Wang et al., 2021) were able to deploy the model in mobile phone devices. In addition, an end-to-end architecture that faces globally the visual inspection process of containers, grouping and solving multiple visual inspection tasks has been presented (Delgado, 2022).

All these methods have in common that the data used is scarce and typically, not public, making the process of training new algorithms complicated because collecting data is a must. Moreover, comparing metrics between other algorithms is next to impossible. Synthetic data generation has proven to be a good approach to tackle the lack of data (Cortes et al., 2022). Data Augmentation is the first step that most of the current algorithms use in order to increment the amount of data with few effort (Zoph et al., 2020; Shin et al., 2020). However, sometimes applying certain distortions to the image is not possible by only using traditional data augmentation and other methods. This has to be taken into account to generate specific transformations that are present in the real target data, so that a high quality and representative amount of data can be provided to be used in training neural network models. A more complex and flexible solution is to generate synthetic datasets based on rendering 3D models or environments from the desired perspective and using a predefined virtual sensor setup. For example, the SUNCG (Song et al., 2017) and the Matterport datasets (Chang et al., 1709) provide synthetic data for indoor scenes along with the corresponding depth and semantic ground truth. In human behaviour and biometric analysis research, synthetic data have an additional advantage in mitigating the difficulties of coping with data privacy laws. Due to the consequent interest, this area has different synthetic datasets. For example, (Wood and Baltrusaitis, 2021) presents a dataset with high realism focused on face data, and the authors in (Erfanian Ebadi et al., 2022) propose a synthetic dataset oriented to people and pose detection. Simulator-based synthetic data generation is gaining popularity too, especially with interactive simulation environments (Nikolenko, 2021), where not only an image with an object is generated but also a complex and auto-labelled scenario with freedom for customization and control over the scenes to be captured. Different environments are proposed for the automotive field, letting the user define the desired sensor setup and interact with other environment agents (Dosovitskiy et al., 2017);(Shah et al., 2018). There are also methodologies for generating synthetic data for training Deep Learning models, where the input data are the 3D assets of the scene and the scene's description file (Aranjuelo, 2021). However, these methodologies are specific to other fields, such as surveillance, and cannot solve our issues. Last, recent generative models, such as StyleGAN3 (Karras et al., 2021), or diffusion models, such as Stable Diffusion (Rombach et al., 2022), might be used for generating high-quality synthetic images. Nonetheless, their training already requires having some sample data and they do not provide the flexible control over the scene that a 3D environment might provide.

Previous studies have established that the integration of synthetic data can augment performance when utilized along-side real-world data. Moreover, related research has furnished evidence supporting the effectiveness of synthetic data in the context of training. Taken together, these findings substantiate the credibility of employing synthetic data during training has the potential to generate noteworthy advancements in computer vision systems (Wood and Baltrusaitis, 2021; Borkman et al., 2021; Jaipuria et al., 2020; Aranjuelo et al., 2022).

## 3. Methodology for labelled dataset generation and annotation

In the context of shipping containers' analysis, there are several visual characteristics that are necessary to reproduce. In this work we tackle the detection and segmentation of damages that the container might have during the shipment, the detection of IMDG markers and the door/no door classification, which we explain in depth in Section 4. In order to tackle these issues, a methodology for synthetic images generation has been implemented as we can see in Fig. 1. The objective of this tagged database is to help researchers train models that are able to learn the location and typology of the previously mentioned elements in a container. In this section, we explain in detail the methodology for the generation of such dataset, specifying the modifications used in the 3D model and the whole pipeline of the methodology, including the annotation process. In addition, we provide publicly the generated dataset containing almost 10,000 images for training and validation and 2480 images for testing, in order to provide open and free data that is scarce in this field. It is important to consider that the presented dataset is entirely generated in a programmatic way. If elements such as damages or a IMDG markers are introduced, we keep track of all of them so we can ensure that the annotation is generated for each item presented.

### 3.1. Classes and annotation formats

In this work, we defined the target classes according to some of the most usual damages we can find in containers and the standardized IMDG markers (International Maritime Organization, 2004). Regarding the different types of damages, several classes have been considered as follows: (i) concave as a component that is damaged by being bowed or bent. This damage is gradual over the length of the component face; (ii) axis is a damage located only on the edges of the container; (iii) dented is defined as a slight hollow in an even surface made by a blow or pressure. Dented walls are generally caused when moving cargo in and out of the container; (iv) finally, perforation is defined as a puncture due to poor handling of forklift, hit by another vehicle or container. In Fig. 2 we can observe the different types of damages.

IMDG (International Maritime Organization and Code, 2020; International Maritime Organization, 2004) codes are a set of visual markers that identify the type of the goods in a container. These visual elements are generated using canonical IMDG markers obtained from the internet. They are classified in nine types of goods regarding the risk they are associated with as shown in Fig. 3.

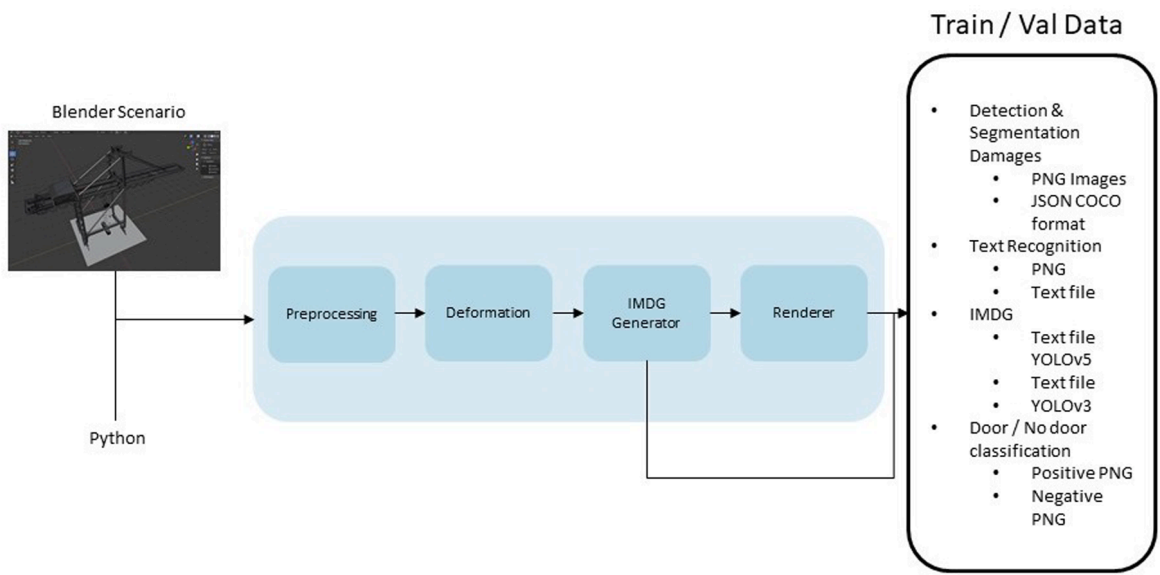

**Fig. 1.** Generation scripts output datasets.

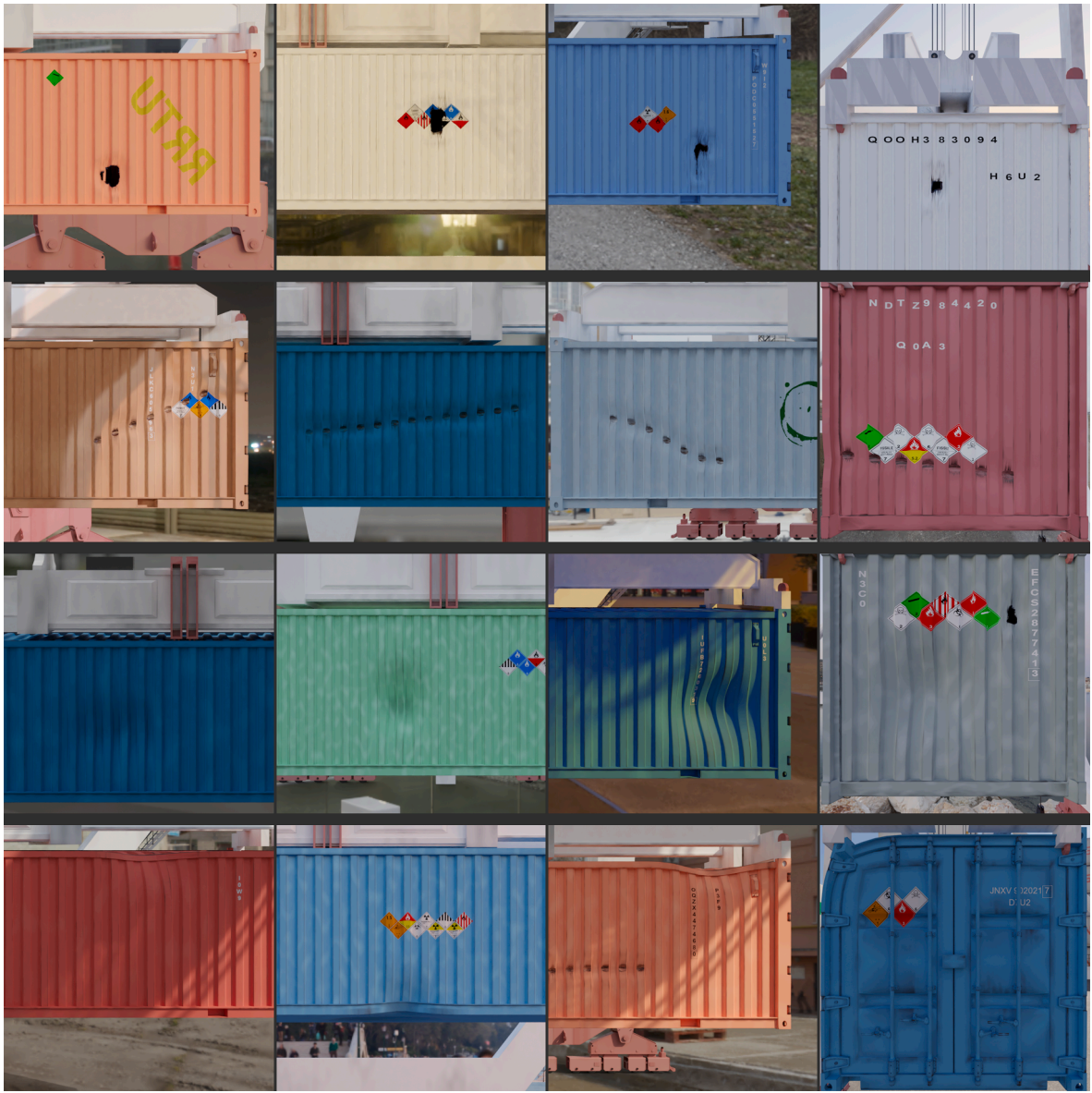

**Fig. 2.** Different types of damages that are generated. From top to bottom we find perforation, dented, concave and axis.

On the other hand, the annotation format depends on the specific task; damage segmentation, IMDG marker detection, text recognition and door/no door classification. Their annotations are deeply linked with the type of visual tasks that are intended to be undertaken. Regarding damage segmentation, we provide a PNG image containing the label of each pixel. In addition, a JSON file is also generated with COCO's format providing the same segmentation information. For the IMDG marker detection task, a YOLO format annotation is provided. The text recognition annotation consists in a folder where there are located all the image crops containing the container identifier and a text file where labels for each crop image are located. Finally, the classification task is organized in two folders separating the positive samples (door images) and the negative samples (no door images).

## 3.2. Scenario and 3D model

For the rendering of container images, Blender 2.91.2 has been used. Cycles has been chosen as render engine due to its greater realism in the results.

The Blender scene is made up of various elements. In addition to the container, an STS crane's model has been introduced. The crane, and especially the clamp that hooks the container, is part of the natural environment where the action captured by the cameras

**Fig. 3.** Types of goods regarding IMDG codes.

happens and provides fidelity to the final render.

There are four virtual cameras placed on each side of the container in the scene, Camera A, Camera B and Cameras Cl and Cr see Fig. 4. Cameras A and B always point at the Back and Front side of the container, however Cameras Cl and Cr point at the door and the no door. The location of the cameras and the specifications are based on the existing setup of the port of Luka Koper. The scene mimics the position, focal length (f) and final resolution to obtain images with similar distortion and properties to the real ones. However, this setup is easily configurable depending on the needs of other similar environments for different use cases.

High Dynamic Range Images (HDRI) are used as sole light source of the scene. This global illumination technique uses an image that is projected to the entire 360° space of the scene. Besides the lighting that the HDRI provides to the scene, the material of the objects will be affected by every pixel of the environment image. The metallic or reflective materials pick up this colour and light information and the final render result will return many nuances. For each new rendered image, different HDRI will be used. This way many visual variations of the same material are obtained. The HDRI is also used as the background of the render. This methodology has the ad of allowing to have a final composition without the need of having a dedicated 3D background scenario. The HDRI chosen as backgrounds for these methodologies correspond to a variety of daytime and nigh-time images and urban or industrial environments with different weather conditions.

The containers have been modelled following the standard ISO measures (Interational Organization for Standardization and Series 1 Freight Containers, 2013), both in dimensions and depth of the undulation. The group of polygons that form the container is subdivided proportionally. This is needed in order to allow the future deformation processes to modify the geometry of the container in a realistic way. In addition, an exhaustive UV texture mapping has been generated in order to correctly place the labels and the text identifiers in relation to container's polygons. This UV texture maps the 3D modelling process of a 3D model's surface to a 2D image for texture mapping. Therefore, the UV texture mapping corresponds to the different container's sides cannot have overlapping polygons and their distribution must be proportionate and homogeneous.

In addition, to have a realistic material we need a group of defects such as dirt, ageing, rust, or scratches. Each material defect is a different layer that is combined with the others to achieve the final texture. In order to create the ageing, regions of the geometry near the corners of the container are found and darkened by a colour ramp. Regarding the dirt, the polygons close to any edge are chosen and painted in light grey. Rust is achieved creating ochre noise over the container surface that mixes with the base colour. Also, this base colour has noise variations in its tone. In addition, a texture with the texts and IMDG markers is placed on the previous result. The container material reflection has been modified in external areas using a mask and introducing a noise variation to make it look older. All these elements form the transformation nodes that we can see in Fig. 5.

### 3.3. Pipeline for data generation

The pipeline defined for the generation of the dataset consists in different modules and are organized as follows: preprocessing, deformation, IMDG marker generator and renderer. The full pipeline can be seen in Fig. 1. These visual aspects could be different in other contexts, but the underlying pipeline would remain the same. The whole pipeline is conducted by a python script which interacts with the Blender scene objects natively. Polygons are modified via python to obtain a visual appearance of the final rendered container that match a real appearance where several images attending to different pre-established points of view will be generated with their corresponding annotation files.

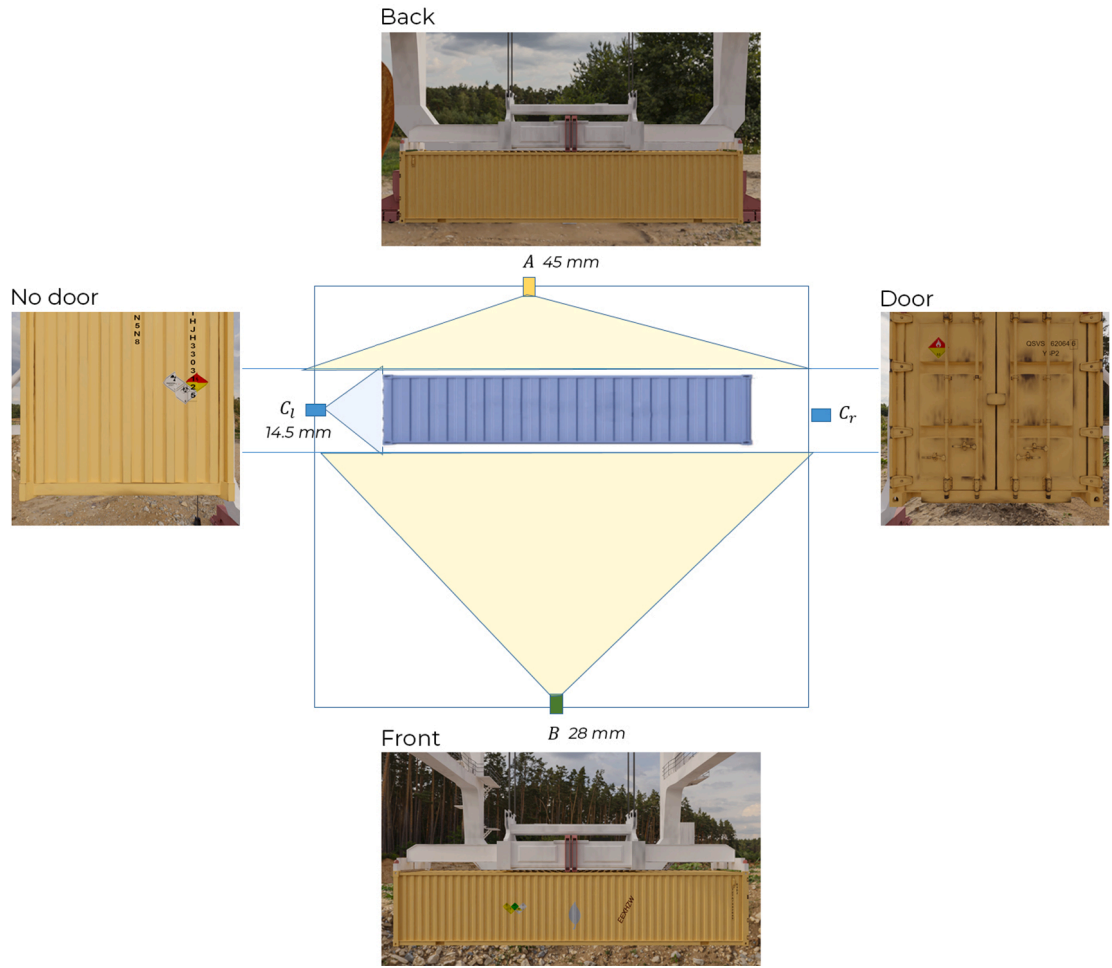

**Fig. 4.** Configuration of the virtual cameras inside the 3D scenario for the SeaFront dataset. Camera A has a focal length of f = 14.5 mm, camera B has f = 28 mm and cameras $C_l$ and $C_r$ have the same focal length of f = 45 mm.

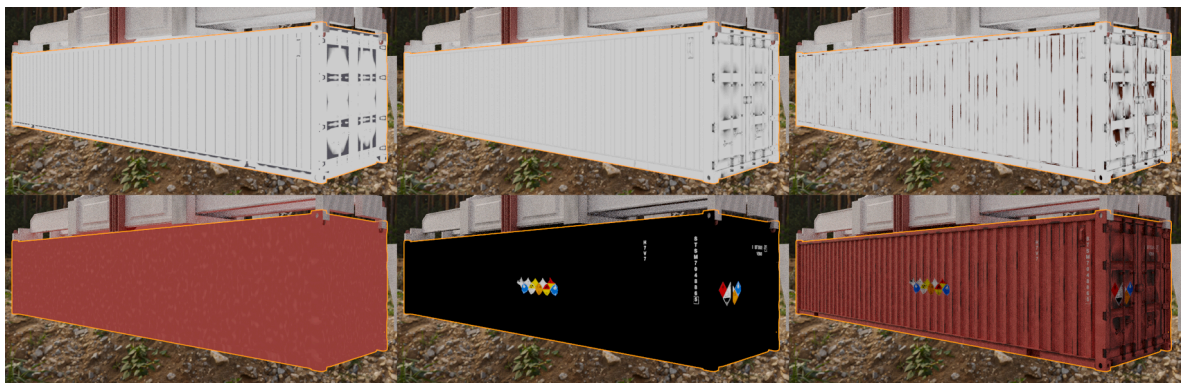

**Fig. 5.** Achieving realistic container representation through material compositing. Top row: Aging, dirt, and rust effects. Bottom row: Tone variation, IMDG markers, and reflections.

#### 3.3.1. Preprocessing

The preprocessing module has multiple tasks. First of all, it sets the framework preferences to enable GPU and CUDA to work. We strongly suggest using a combination of CPU and GPU as it will speed up the rendering task greatly. Secondly, it changes randomly the colour of the container's material given a list of RGB values. Doing it this way, we can add or remove colours without further issues. In addition, it changes specific material's parameters that ages the container. This is done due to the fact that a realistic container is needed. Otherwise, we would have multiple containers that has the same ageing material. The range of the values were set empirically and then the values are selected randomly through the pipeline. Next, we select randomly a background for the scene among all that has been preprocessed beforehand. We use public free HDRI files. Finally, we define a translation range for the bottom and the upper cameras. This creates a wider range of diversity in the dataset, where the containers might be seen from multiple perspectives.

#### 3.3.2. Deformations

The deformation module aims reproduce and combine the various types of deformation. Variations occur both in the location, as well as in its shape and depth. For each one of the deformations already mentioned, a different modelling technique has been used. In all cases, the starting point is a container without any damage and a combination of these is added randomly. The size, shape and intensity of each damage is controlled, as well as the location. Thus, we set a probability of a 40% for the pipeline to generate a deformation. This feature promotes the dataset to be more balanced between containers with and without. We aim to have a 60% of the samples without any of them. If no deformations are chosen, all the structures and objects will be deleted except for the container. On the other hand, if deformations are chosen, the pipeline generates four main damages; axis, concave, dented or perforation. The type and the number of deformations is also uniformly randomly selected.

The axis and concave deformations are really similar in terms of how are generated. These deformations are created by a lattice that wraps around the surface of the object and distributes its subdivisions proportionally to it. The main difference between both of them are the location of the damage. While axis its produced on the edges of the container, the concave is generated on the surface of the different faces. When a deformation is generated, a group of vertices of the area to be deformed is chosen and their position is moved along the corresponding axis, within a realistic range. In addition, the translation on the selected vertices produces a smooth dragging effect on the neighbouring vertices and thus a progressive sinking of the surface is achieved. Moreover, comparing the modified container with a copy of the original one, we are able to extract a new object that will be rendered as the annotation of the damage.

Regarding dents, they are a more subtle type of deformation. It is characterized by a fine, elongated fold on the surface. To achieve the dented effect in the face of the container, a boolean 'Intersect' function in Blender is used to remove parts of the original 3D model with a row of wedge-shaped objects located on the external surface of a container's face. These objects are placed on a spline that alternates between linear and sinusoidal curves. To control the position in the metal folds, a lattice is created that coincides with them and, knowing the position of these vertices, the pieces that will generate the dented object are placed and used by the boolean 'Intersect' function. Additionally, this lattice uses the position of those vertices to slightly translate and sink the polygons on the container's surface for a more realistic feel. In addition, a type of aged material is transferred to the sunken area. On the other side, perforations work similar to dents. In order to generate holes in the faces of the container, it is necessary to extract and deform a set of polygons from the 3D model. A boolean function is also used on the surface of the container from a variety of 3D objects that have a round and irregular shape. The deformed polygons and those in its contour are selected and their position is translated into the container in a smooth manner. As a final step, aged material is also transferred to the edges of the perforated area.

#### 3.3.3. IMDG marker generator

Another main process is the IMDG marker generator. IMDG markers are added through the UV texture image see Algorithm A1. The annotation process of these IMDG markers involves different steps. First n-randomly selected IMDG markers are placed in the UV texture image in a random position following a two rows rhombus pattern see Fig. 6. A different unique UV texture mask is generated, for each one of these selected markers. These UV texture masks are binary. Once an UV texture mask is generated, see Algorithm A2, they are applied one at a time to the container in the Blender scene and a rendered image is generated. The resulting image is a binary

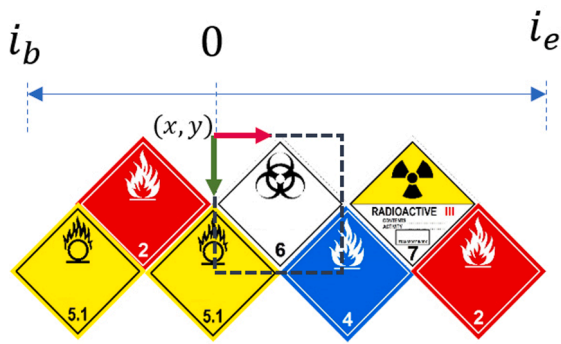

**Fig. 6.** Location of the IMDG markers along the UV texture.

image mask that will be used to locate the position of the IMDG marker in the final image.

---

**A1 ProcessUVTexture:** An algorithm to generate an UV image with IMDG markers and identification text in it.

    **input** : An UV Image $I_{uv}$ of size $w \times l$
    **output**: $I_{uv}$ modified and a bunch of masks for each one of the IMDG markers and texts

*Overall processes that apply additional noise*
**foreach** *face* $f \in \{door, no\ door\}$ **do**
    **if** GetRandom(*Number*) $< th_{\alpha,f}$ **then**
        DrawBrand($f, I_{uv}$,GetRandom(*brand*))
    **if** GetRandom(*Number*) $< th_{\beta,f}$ **then**
        DrawMainText($f, I_{uv}$,GetRandom(*Text*))

*Aggregate IMDG markers and Identification Texts to the UV texture*
**foreach** *face* $f \in door, no\ door, back, front$ **do**
    $imdg\_list \leftarrow$ GetRandom(*IMDGs*)
    CreateIMDGMasks($f, I_{uv}, imdg\_list$) $\rightarrow$ *IMDG Masks and IMDG UV Mask Textures*

    $text\_list \leftarrow$ GetRandom(*Texts*)
    CreateTextMasks($f, I_{uv}, text\_list$) $\rightarrow$ *Text Masks and Text UV Mask Textures*
SaveToDisk($I_{uv}$)

---

IMDG markers initial location is calculated and are distributed in the following way. First, $i_b$ and $i_e$ are calculated, being N the number of IMDG markers to insert, $i_e = N/2 + 1$ and $i_b = i_e - N$. Secondly, the algorithm iterates from $i_b$ to $i_e$ placing each of the IMDG markers into the UV texture image. The location of each marker is computed as follows:

$$x_i = \lceil x + \frac{i*w}{2} \rceil$$

$$y_i = \begin{cases} \lceil y \rceil & if\ |i|\%2 == 0 \\ \lceil y + \frac{h}{2} \rceil & otherwise \end{cases}$$

After this calculation, for each marker several images are generated. On one hand, the marker is drawn in the container's general UV texture image. On the other hand, an UV texture mask is also generated only for that marker. This UV mask is then applied to the scene and used to render a new image mask for that marker. A visual explanation can be found in Fig. 7.

A connected component analysis will be performed in this last rendered image to obtain the bounding box for that marker and finally, this bounding box will be saved in a file as an annotation.

### 3.3.4. Renderer

The renderer module loads the texture generated by the IMDG marker generator and starts setting up parameters. Using the Cycles render engine from Blender, it generates a realistic image for each camera in a PNG format. These images are organized also regarding their task. Then, an image for each camera is also rendered and assigns a semantic category to each pixel of the images. In order to recover the class of the segmented object, it is needed to load the PNG and compute the integer value of class = $Image_{i,j}/100$. For the identifier of the instance can be calculated as instance = $Image_{i,j}\%100$. Unlabelled pixels (void) are assigned a value of 0.

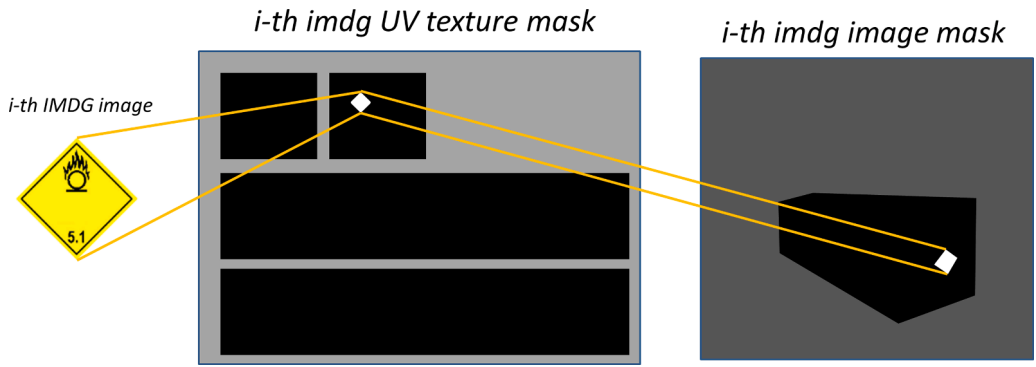

**Fig. 7.** Mask generated for each IMDG marker and text.

---

**A2 GenerateIMDGMasks:** An algorithm to incorporate IMDG markers to an UV texture and to generate all the information required to further calculate the bounding boxes.

**input** : An UV Image $I_{uv}$ of size $w \times l$
**input** : A $(x, y)$ location for the UV markers
**input** : The number of markers N
**input** : A list $imdg\_list$ with N random IMDG images
**output:** $I_{uv}$ modified with the markers incorporated and a bunch of masks, one for each one of the IMDG markers

$i_e \leftarrow \lfloor \frac{N}{2} + 1 \rfloor$
$i_b \leftarrow N - i_e$
**for** $i \leftarrow i_b$ **to** $i_e$ **do**
    *for each IMDG marker i in the imdg_list generate its output*
    $w, h \leftarrow IMDG_{i-i_b}.shape$
    $x_i \leftarrow ceil(x + i * w * 0.5)$
    **if** $i \bmod 2 == 0$ **then**
        $y_i \leftarrow ceil(i)$
    **else**
        $y_i \leftarrow ceil(y + i * h * 0.5)$
    *Draw marker in the overall $I_{uv}$ image*
    DrawIMDG($IMDG_{i-i_b}, I_{uv}$)
    *Draw marker in the local $UVMask_i$ image*
    DrawIMDG($IMDG_{i-i_b}, UVMask_i$)
    *Apply local $UVMask_i$ to the container*
    ApplyUV($UVMask_i$)
    *Render scene with current UV applied*
    $Mask_i \leftarrow Render()$
    SaveToFile($Mask_i$)

---

Finally, the segmentation annotator provides annotations for each deformation in YOLO and COCO format.

### 3.4. Statistics

In this section, we provide statistical analysis about the generated dataset. We show some aspects on the distribution of classes and their usual bounding box size. The dataset is split in training and validation and consist in a total of 9888 images where 7910 are for training and 1978 for validation. In order to provide results, a test dataset has also been generated with 2480 images.

Regarding detection and segmentation of damages, our goal is to achieve a pipeline which can generate balanced data. Thus, we prove this as we can see in Table 1. We aimed to generate for each dataset a 60% of the samples without any damages and a 40% with damages. We found that this division is enough for current state of the art models to not overfit on their respective classes and have a wide variety of representatives between containers and containers with deformations.

The distribution of the different deformation classes can be seen in Table 2. The class container is the predominant because a container is necessary to present damages. However, the other classes that represent the different types of damages are balanced uniformly. This is due to randomly selecting the class and the number of different deformations in a single container.

Regarding the distribution for the IMDG markers can be found in Table 3. All the IMDG markers are uniformly balanced to get better results in the training phase. However, while a given IMDG class can appear in some of the images, the container class will also appear in all the images, as stated previously from the other visual task. Regarding the class text, it will always appear unless there is some noticeable occlusion. In addition, more than one text can appear in a single image and that is the reason why the classes text and container are unbalanced in nature.

**Table 1**
Frequency distribution of the different dataset's partitions whether a container has damages or not.

|           | Train | Val  | Test |
|-----------|-------|------|------|
| No damage | 5563  | 1406 | 1821 |
| Damaged   | 2347  | 572  | 659  |

**Table 2**
Distribution of the different classes for the visual task of segmentation and detection of damages.

|  | Train | Val | Test |
|---|---|---|---|
| Container | 7910 | 1978 | 2480 |
| Axis | 1081 | 259 | 245 |
| Concave | 636 | 171 | 182 |
| Dented | 796 | 186 | 235 |
| Perforation | 775 | 189 | 246 |

**Table 3**
Distribution of the different classes for the visual task of IMDG detection.

|  | Train | Val | Test |
|---|---|---|---|
| Text | 11,825 | 2932 | 3655 |
| C1.1 | 1026 | 253 | 319 |
| C1.2 | 1031 | 271 | 304 |
| C1.3 | 1002 | 250 | 341 |
| C1.4 | 1067 | 250 | 333 |
| C2.1 | 1050 | 257 | 324 |
| C2.2 | 961 | 270 | 307 |
| C2.3 | 1100 | 258 | 310 |
| C2.4 | 1014 | 276 | 313 |
| C2.5 | 1082 | 239 | 333 |
| C3.1 | 1011 | 248 | 300 |
| C3.2 | 1020 | 289 | 331 |
| C4.1 | 1052 | 252 | 312 |
| C4.2 | 1078 | 252 | 319 |
| C4.3 | 993 | 259 | 325 |
| C4.4 | 962 | 271 | 320 |
| C5.1 | 983 | 249 | 339 |
| C5.2 | 1031 | 263 | 353 |
| C5.3 | 1040 | 267 | 311 |
| C6.1 | 1060 | 278 | 346 |
| C6.2 | 1099 | 250 | 318 |
| C7.1 | 983 | 255 | 337 |
| C7.2 | 1042 | 232 | 302 |
| C7.3 | 1054 | 257 | 349 |
| C7.4 | 1002 | 247 | 331 |
| C8.1 | 1065 | 255 | 314 |
| C9.1 | 1065 | 250 | 323 |
| Container | 7910 | 1978 | 2480 |

Finally, we ensured that the pipeline generates a variety of objects that have different sizes and shapes in order to provide more diversity within the dataset, as can be seen in Fig. 8.

For the door/no door classification a dataset from cameras $C_l$ and $C_r$ is used for training and validation, and it is balanced. For testing we have done different experiments. In some cases, we have used cameras A and B apart from the other two and in those cases, the dataset becomes unbalanced because the number of negatives increases while the same number of positives is maintained. This problem, however, is not so critical as it can be alleviated by using dataset balancing techniques such as applying undersampling in the most populated class.

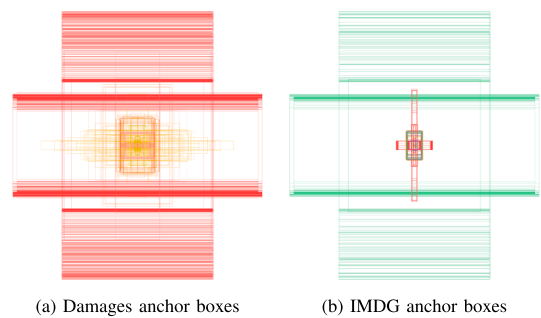

(a) Damages anchor boxes     (b) IMDG anchor boxes

**Fig. 8.** Anchor boxes from the training dataset for the detection of different visual tasks.

## 4. Definition of visual task for the methodology's validation

In this section, we propose and define three visual tasks that are found recurrently in the state of the art and aims to automatize the manual inspection of the containers. They also provide a way to validate the dataset generated through the proposed methodology.

### 4.1. Damage segmentation and detection

This task involves predicting the location and the label of the container and its four possible deformations labelled as axis, concave, dented and perforation. The location of these five classes can be done via detection using bounding boxes, where the output should be a confidence value and their coordinates. In addition, it can also be done via semantic segmentation predicting a class per-pixel of the image without considering the different instances of the same class. Both approaches are evaluated using COCO metrics (Lin et al., 2014). Firstly, it reports the average precision (AP) for an intersection over the union.

(IOU) threshold of 0.5 ($AP_{50}$), 0.75 ($AP_{75}$) and 0.5:0.05:0.95 (AP) which calculates the AP from 0.5 to 0.95 with steps of 0.05. It also reports the average recall (AR) for a range of thresholds of 0.5:0.05:0.95. It also provides these metrics for all, small ($AP_S$, $AR_S$), medium ($AP_M$, $AR_M$) and large objects ($AP_L$, $AR_L$). The IoU for detection is calculated using the overlap of the detection's bounding box with the ground-truth while the segmentation is defined as the correct pixels divided by the annotated pixels.

### 4.2. IMDG detection

This visual task aims to detect different elements in the image. As mentioned before, these classes that make up the output of the detector are container, text (vertical or horizontal) and all the typologies of IMDG markers. The detection of the container obeys to the necessity of reducing the area of the image to be analysed. So, this detection will be used in further steps to filter the image and to reduce the computational cost of the rest of the modules involved in the container inspection. In the same way the detection of text is carried out to reinforce the output of a subsequent text detector. Images used to train this model has been generated by the scripts previously described. The metrics used to evaluate this task are similar to the damage segmentation and detection task. We report the average precision (AP) for an intersection over the union (IOU) threshold of 0.5 ($AP_{50}$) and 0.5:0.05:0.95 (AP). In addition, we also use precision (P) and recall (R) metrics for class with an IOU of 0.5.

### 4.3. Door/no door classification

This task performs a binary classification to detect whether a door from a container is present or not. When the classifier detects a container door, 1 is returned and when not, 0 is returned. The output of the classifier is the prediction label and its confidence value. Moreover, in order to evaluate the classifier, the precision and recall metrics have been used.

The training of the classifier has been done using a subset of the synthetic images, where only Door and No door images have been used. The positive samples are the ones which contain a door in camera $C_r$. On the other side, the negative samples are from camera $C_l$ and do not contain a door, see Fig. 4.

We carried out two main experiments. A first experiment uses a subset of 1240 images from cameras $C_l$ and $C_r$ as we did during training. A second experiment has been carried out using all the images (2480) from all the different cameras, $C_l$, $C_r$, A and B. Images from A and B cameras are considered as negatives.

## 5. Experiments and results

In this section we show that training a model with accurately generated synthetic images can lead to competitive results, positioning it as a viable alternative to real datasets that are difficult to obtain because of private entities in port environments and the difficulty to annotate them. Thus, these experiments aim to validate our proposed methodology to generate synthetic labelled datasets. For training and evaluation, we used a single NVIDIA Tesla V100-SXM2 of 32 GB.

### 5.1. Damage segmentation and detection

Regarding damage segmentation and damage detection, we studied how the proposed synthetic dataset works with the various state-of-the-art models. To do this, we have trained and tested models that only do detection, other that only do segmentation, and other can that do both tasks together. Therefore, to tackle detection and segmentation all together we decided to use the two-stage detector R-CNN family. We used Mask R-CNN (He et al., 2017) that extends the Faster R-CNNN (Ren et al., 2017) by adding a segmentation branch in parallel to the existing detection branch during training. Cascade Mask R-CNN (CM R-CNN) (Cai and Vasconcelos, 2019) was used as a hybrid between Cascade R-CNN and Mask R-CNN where we use the power of cascade regression for detection and the mask for segmentation. Mask Scoring R-CNN (MS R-CNN) (Huang et al., 2019) was also used as an extension of Mask R-CNN with MaskIoU Head, which takes the instance feature and the predicted mask together as input, and predicts the intersection over union between input mask and ground truth mask. In respect to only detection, we used the previously mentioned Cascade R-CNN which is a multi-stage extension of the Faster R-CNN, where detector stages deeper into the cascade are sequentially more selective against close false positives. In addition, we trained Dynamic R-CNN (Zhang et al., 2020) which adjusts the label assignment criteria and the shape of regression loss func-tion automatically based on the statistics of proposals during training. Cascade RPN Faster R-CNN (Vu et al., 2019) (CRPN Faster R-CNN) was also trained. This network performs multi-stage anchor refinement and uses adaptive convolution. Besides, we trained Sparse R-CNN (Sun et al., 2011) where object candidates are given with a fixed small set of learnable bounding boxes. Finally, we used Faster R-CNN to compare it with all previous methods. In addition, only segmentation has been tackled using SOLO (Wang et al., 2020) which assigns categories to each pixel within an instance according to the instance's location and size and SOLOv2 (Wang et al., 2020) which dynamically learns the mask head of the object segmenter. All the models were trained for 20 epochs with a decreasing the learning rate of 0.01 at 16 and 19 epochs. We used the MMDetection framework (Chen et al., 1906) in Pytorch, with a batch size of 16 samples with an input image of 1333x800. We used the test dataset to provide results for detection that can be found in Tables 4 and 5. As we can see, Cascade Mask R-CNN has the best AP results. Specially, we have a solid precision for large and medium objects. However, the smaller it gets the worse the precision. That is where Dynamic R-CNN outperforms other networks in small objects, as due to the dynamic label assignment component of its network is able to improve the quality of the proposal as the training goes. Nevertheless, the values are low for all the networks, and this is normal as deformations can be sometimes small and difficult to locate by the detector, especially after resizing the original image to the input network size. Cascade R-CNN, which is focused only on detection, provide similar results to Cascade.

Mask R-CNN getting better results in large objects. However, we have to take into consideration that Cascade Mask R-CNN is able to solve detection and segmentation tasks at once. Recall performance also affected by small objects but provide good results in other sizes. We can observe the same model behaviour seen in the average precision results. At the end, we provide a good baseline and starting point with a solid average precision and average recall for different models in the state of the art.

Regarding segmentation, results are found in Table 6 and Table 7. We can see that we get similar result to the detection task with the same insights. Cascade Mask R-CNN provides the best results except for the smallest objects, where Mask Scoring R-CNN does. Thus, we provide a solid average precision and recall for the segmentation task with multiple models in the state of the art.

*5.2. IMDG detection*

For the IMDG detection task, we studied how the most used detector family YOLO behaves with the proposed synthetic dataset and we checked if it is able to solve the current task. In Table 8 we can see the different models used to tackle the detection of the different markers in the container. Due to YOLOv7 (Wang et al., 2022) architecture and its improvements in its backbone, we have a better representation and classification of the different objects to detects. That is why we have the best results in precision and $mAP_{[0.5:0.95]}$. However, YOLOR (Wang et al., 2105) is a model that does not use anchors and it gets a better recall metrics trading off with precision. Nevertheless, we believe that YOLOv7 is the one that has the most balanced performance along the evaluation metrics, and it has proven to be a good choice with a balanced trade-off between computation cost and performance. In addition, the YOLO can be used in several architecture depths, from tiny to extralarge architecture, which differ mainly in the number of parameters and operations that are directly connected to the size of the model and the performance. In this studied we focus only YOLO architectures that have similar complexities. Among them, YOLOv7 model was selected as it has enough representational capability to tackle this task in the presented dataset.

In Fig. 9, we can see an example of the resulting inference of the detection and the segmentation using Cascade Mask R-CNN. The detection and the segmentation work well for the synthetic test dataset. Even though, sometimes the segmentation can be less accurate for damages, the detections are surrounded by a fairly precise bounding boxes.

YOLOv7 architecture is an extension of YOLOv4 (Bochkovskiy et al., 2020), Scaled-Yolov4 (Wang et al., 2021) and YOLOR by adding several architectural reforms as an Extended Efficient Layer Aggregation Network (E-ELAN) and a trainable bag of freebies. E-ELAN is the computational block in the YOLOv7 backbone, and it is designed to improve speed and accuracy. On the other hand, the bag of freebies is a bunch of methods to increase the performance of the model. As mentioned before detection classes has been extended to container and text as well. We trained the model for 150 epochs with learning rate of 0.01. We use Wang's YOLOv7 Pytorch framework (Wang et al., 2022) with a batch size of 16 and an input image resolution of 640x640 letterbox scaled.

As expected, results in Table 9 highlights that some classes are easier to detect than others. Several classes present more inter-class correlation than others and therefore they are harder to differentiate. This entails that the precision of similar classes might be lower. In contrast, other classes have more pronounced and robust visual features and are easily learned by the detector such as the container and the text classes.

A further in-depth analysis on the most complicated classes could offer new lines of improvement in the precision obtained for these classes. The overall precision of the detector achieves a precision of 79.9 and a recall of 95.3 with a mAP50 of 83.0. Although obtained results can still be improved, from the authors' perspective these results offer a very promising way to evolve the model using synthetic data.

**Table 4**
Average precision of detection of damages.

| Method | AP | $AP_{50}$ | $AP_{75}$ | $AP_S$ | $AP_M$ | $AP_L$ |
| --- | --- | --- | --- | --- | --- | --- |
| CM R-CNN (Cai and Vasconcelos, 2019) | **71.5** | 89.9 | 79.2 | 12.5 | **50.3** | 77.8 |
| Mask R-CNN (He et al., 2017) | 67.5 | 89.2 | 75.5 | 12.4 | 45.9 | 73.0 |
| MS R-CNN (Huang et al., 2019) | 69.0 | 89.6 | 76.6 | 14.7 | 47.8 | 74.8 |
| Faster R-CNN (Ren et al., 2017) | 69.1 | 90.5 | 78.4 | 11.4 | 47.1 | 75.3 |
| CRPN Faster R-CNN (Vu et al., 2019) | 63.4 | 85.1 | 69.6 | 11.6 | 37.9 | 70.4 |
| Dynamic R-CNN (Zhang et al., 2020) | 68.4 | **91.5** | 77.9 | **18.7** | 46.1 | 74.0 |
| Cascade R-CNN (Vu et al., 2019) | 71.4 | 90.4 | **79.8** | 11.9 | 47.7 | **78.8** |
| Sparse R-CNN (Sun et al., 2011) | 65.7 | 89.6 | 71.2 | 18.2 | 40.7 | 74.7 |

**Table 5**
Average recall of detection of damages.

| Method | **AR** | **AR$_S$** | **AR$_M$** | AR$_L$ |
|---|---|---|---|---|
| CM R-CNN (Cai and Vasconcelos, 2019) | **74.7** | 13.8 | **53.9** | 80.9 |
| Mask R-CNN (He and Gkioxari, 2017) | 71.7 | 13.2 | 50.9 | 77.7 |
| MS R-CNN (Huang et al., 2019) | 72.9 | 15.1 | 51.7 | 78.9 |
| Faster R-CNN (Ren et al., 2017) | 72.7 | 12.5 | 52.0 | 79.1 |
| CRPN Faster R-CNN(Vu et al., 2019) | 69.5 | 15.2 | 47.7 | 75.9 |
| Dynamic R-CNN (Zhang et al., 2020) | 72.6 | **19.6** | 51.7 | 78.3 |
| Cascade R-CNN (Vu et al., 2019) | 74.6 | 12.8 | 52.5 | **82.0** |
| Sparse R-CNN (Sun et al., 2011) | 75.8 | 25.0 | 55.7 | 81.2 |

**Table 6**
Average precision of segmentation of damages.

| Method | AP | AP50 | **AP75** | AP$_S$ | AP$_M$ | AP$_L$ |
|---|---|---|---|---|---|---|
| CM R-CNN (Cai and Vasconcelos, 2019) | **67.5** | **88.6** | **74.4** | 12.6 | **45.8** | **73.3** |
| Mask R-CNN (He and Gkioxari, 2017) | 55.2 | 71.5 | 61.0 | 12.0 | 31.0 | 59.9 |
| MS R-CNN (Huang et al., 2019) | 58.3 | 73.6 | 64.6 | **13.6** | 37.1 | 62.1 |
| SOLO (Wang et al., 2020) | 46.3 | 63.6 | 48.4 | 6.0 | 22.3 | 52.2 |
| SOLOv2 (Wang et al., 2020) | 52.1 | 70.2 | 56.4 | 7.2 | 26.6 | 58.1 |

**Table 7**
Average recall of segmentation of damages.

| Method | AR | AR$_S$ | AR$_M$ | AR$_L$ |
|---|---|---|---|---|
| CM R-CNN (Cai and Vasconcelos, 2019) | **71.1** | 14.1 | **49.4** | **77.0** |
| Mask R-CNN (He and Gkioxari, 2017) | 59.3 | 18.4 | 36.5 | 63.0 |
| MS R-CNN (Huang et al., 2019) | 61.1 | **19.4** | 39.6 | 64.4 |
| SOLO (Wang et al., 2020) | 51.5 | 9.5 | 27.9 | 57.0 |
| SOLOv2 (Wang et al., 2020) | 56.8 | 11.4 | 34.2 | 61.6 |

**Table 8**
Comparison between other state-of-the-art detection models.

| Method | P | R | mAP50 | mAP |
|---|---|---|---|---|
| YOLOv5-L (Jocher et al., 2022) | 75.4 | 89.8 | 79.2 | 73.4 |
| YOLOv6-S (Li et al., 2209) | 78.0 | 93.0 | **84.0** | 79.0 |
| YOLOv7 (Wang et al., 2022) | **79.9** | 95.3 | 83.0 | **79.6** |
| YOLOR-P6 (Wang et al., 2105) | 51.2 | **98.6** | 77.4 | 72.2 |
| Scaled-YOLOv4-P7 (Wang et al., 2021) | 56.9 | 96.3 | 83.7 | 79.4 |

### 5.3. Door/no door classification

The door/no door classification has been done using ResNet-50 We trained the model for 100 epochs with the proposed dataset.

As mentioned before, we have done two different experiments. One experiment has been carried out using a subset from cameras $C_l$ and $C_r$, similar to the dataset used for training. The results we have obtained can be found in Table 10. We got a precision and a recall of 1.0 since no false negatives or false positives have been detected.

The second experiment has been made using all the images from the dataset containing more negatives, using the camera $C_r$ for the positive images and $C_l$, A and B cameras for the negative images. There are 2480 images of which 620 contain a door and the other 1860 do not. The results of this test can be found in Table 10. We can see that having added some negatives not seen by the classifier to the dataset, some false positives have appeared, but still precision and recall have a high value.

## 6. Domain gap

When a model is trained with a source domain and tested on a target domain, it exists a difference between both data distributions that degrades the model accuracy. The difference between both domains is known as domain gap. The use of synthetic data often maximizes this gap. Domain adaptation techniques are required if the model does not generalize well enough in the target domain.

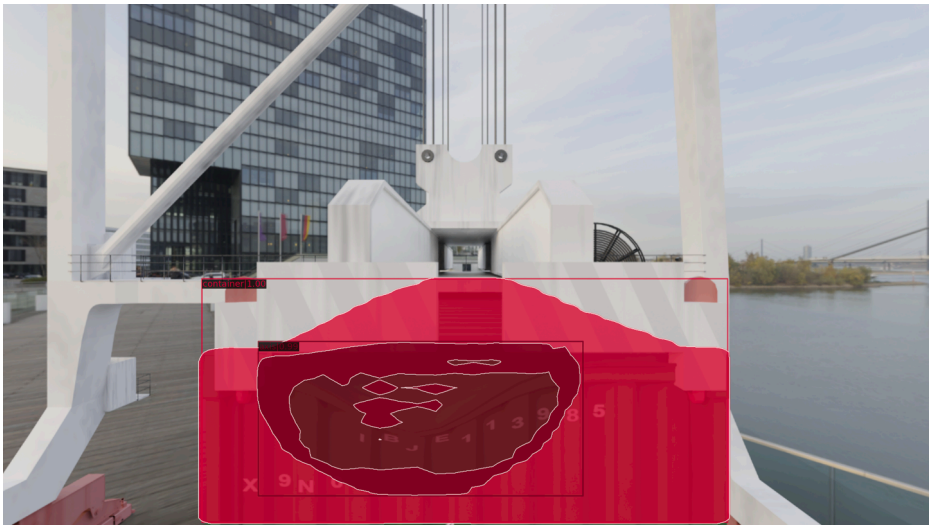

**Fig. 9.** Detection and segmentation of a synthetic generated image.

**Table 9**
Metrics for each class from YOLOv7.

| Class | P | R | $AP_{50}$ | AP |
|---|---|---|---|---|
| text | 96.0 | 86.7 | 93.0 | 79.4 |
| C1.1 | 97.5 | 97.2 | 99.3 | 96.0 |
| C1.2 | 52.6 | 88.5 | 82.2 | 79.5 |
| C1.3 | 64.4 | 63.5 | 78.2 | 75.8 |
| C1.4 | 63.8 | 64.3 | 76.2 | 74.3 |
| C2.1 | 51.1 | 98.1 | 53.1 | 50.8 |
| C2.2 | 48.0 | 98.7 | 50.7 | 48.6 |
| C2.3 | 99.2 | 99.7 | 99.6 | 96.0 |
| C2.4 | 99.0 | 98.1 | 98.7 | 95.2 |
| C2.5 | 49.3 | 99.2 | 53.5 | 51.0 |
| C3.1 | 48.1 | 98.3 | 49.4 | 46.9 |
| C3.2 | 51.7 | 98.5 | 55.4 | 52.8 |
| C4.1 | 99.7 | 99.4 | 99.6 | 95.9 |
| C4.2 | 99.0 | 98.1 | 98.8 | 95.3 |
| C4.3 | 98.4 | 99.4 | 99.6 | 95.6 |
| C4.4 | 99.4 | 98.8 | 99.6 | 95.8 |
| C5.1 | 98.5 | 99.7 | 99.6 | 96.7 |
| C5.2 | 98.8 | 97.7 | 99.5 | 96.4 |
| C5.3 | 98.5 | 99.4 | 99.7 | 96.5 |
| C6.1 | 50.8 | 97.7 | 51.7 | 49.3 |
| C6.2 | 98.8 | 97.8 | 99.1 | 95.6 |
| C7.1 | 99.4 | 99.4 | 99.6 | 95.8 |
| C7.2 | 47.6 | 98.3 | 50.0 | 47.4 |
| C7.3 | 98.2 | 98.6 | 99.5 | 95.8 |
| C7.4 | 52.5 | 98.8 | 55.6 | 52.8 |
| C8.1 | 99.7 | 99.4 | 99.6 | 96.4 |
| C9.1 | 98.9 | 97.8 | 98.7 | 94.6 |
| container | 100 | 100 | 99.9 | 99.9 |

**Table 10**
Precision and recall of door classification.

| | P | R |
|---|---|---|
| Experiment 1 | 100 | 100 |
| Experiment 2 | 90.4 | 96.0 |

It is important to evaluate if the proposed pipeline is good enough to overcome the domain gap or we need to use domain adaptation techniques in order to make it work. Nevertheless, acquiring real tagged data in this field is an arduous challenge considering the fact that there are several privacy and confidentiality constraints in ports. In addition, there is a lot of time and cost that it is needed to devote in the data collection process which makes applying and comparing state-of-the art techniques a difficult task. This is especially difficult for damages, as there is an unbalance in nature between containers that have damages and those that not.

Therefore, we present some limitation in our analytic evaluation. Firstly, it is not feasible to evaluate all the IMDG marker classes presented due to the extreme rarity of some markers. For example, IMDG marker C7.3 which we can see in Fig. 3, is part of the radioactive markers family and identifies fissile materials such as Uranium-235, Plutonium-235, Thorium-232, and others that may be transported. This makes it nearly impossible to evaluate using real images due to stringent security protocols and measures. However, we opted to address this issue by conducting evaluations on a restricted subset of classes and images. This approach should provide sufficient insights to extrapolate the findings to additional classes. On the other hand, similar to what has been stated previously, it is difficult to build a representative real dataset of damage detection and segmentation as its data is very scarce. Containers with damages represent a small population of all the containers that goes through in a harbour, being a total of less than 1% out of in a year. However, we have collected and evaluated empirically a small set of unlabelled images some of which are from a different set up. For the rest of the tasks, we have labelled test set of real container images from Luka Koper while using the STS crane. This dataset comprises 1000 distinct images of 1000 diverse containers. These images were captured under varying environmental conditions, including changes in lighting and weather, on different days. To ensure uniformity across the dataset, we have distributed a 25% representation of each camera. In addition, we introduced a modification to our dataset by incorporating a photorealistic montage created from 20 real images. The purpose of this modification was to include a small subset of IMDG markers belonging to class C4.3.

In damage segmentation and detection, we are able to detect and segment the containers in all the different views of the STS crane. However, damages have a rougher time being detected as those can manifest as really small deformations or multiple different variations. Despite that, the model trained with our proposed method is able to detect and segment some of the damages even with another camera setup as we can see in Fig. 10.

For the IMDG detection task, we firstly analyse the text class and we get a precision of 52.2% and a recall 40.2%. Regarding the container, we get a precision of 95.3% and a recall 88.2%. As we can see, analytically the results have dropped a bit from the synthetic counterpart. This is expected as we are using a different domain of images. However, the container detection performs really solid with real data, detecting most of the annotated containers. It has some struggles detecting multiple containers that are next to each other and some false positives arise by detecting containers in the background which are not annotated. Nevertheless, both of these cases are not contemplated within the synthetic training data and that is why it might lead to errors. We have observed that the text class exhibits the poorest performance based on its evaluation metrics form the synthetic results. This can be attributed to two primary factors. Firstly, the model detects correctly the vertical texts and the horizontal orientation in chunks, resulting in multiple false positives and potential false positives that could be resolved by aggregating bounding boxes. Additionally, the model identifies text that is irrelevant to the ID while failing to detect text that is presented in alternative shapes or is occluded or too small. We hypothesize that this is due to slight variations between the real-world text and the synthetic text used for training the model. We postulate that the observed discrepancy in model performance between real-world and synthetic text is attributable to subtle variations inherent to the two data domains. Specifically, we suggest that the synthetic text used for training the model does not fully capture the diverse range of visual characterstcis present in real-world text. This hypothesis is corroborated by our empirical findings, which demonstrate that the model's container detection ability exhibits a higher degree of similarity between synthetic and real-world data. Thus, there is still scope for reducing the domain gap in the pipeline with respect to the container ID text, such as by incorporating different fonts, text sizes, and colors in the training dataset. Nonetheless, certain cases are not accounted for in the synthetic training data, which could lead to errors. Hence, it is imperative to define these cases as much as possible in the pipeline, enabling the model to generalize across various scenarios. Results can be seen in the Fig. 11. Regarding the detection of the IMDG marker, our evaluation yielded a precision score of 99.7% and a recall score of 100%, with a mean average precision of 99.5%. These outcomes were achieved within a specified class of images from a limited subset captured by cameras Cl and Cr. Our observations demonstrate that these cameras perform exceptionally well in detecting the specified class, providing evidence that analogous classes may be extrapolated to these cameras. Although the evaluation of other classes may not yield identical outcomes, our study suggests the potential for certain classes to perform well under similar conditions. In Fig. 12 we can see the different results.

Finally, we conducted two experiments to evaluate the efficacy of our door/no door classification model in overcoming the domain gap. To do so, we followed the same evaluation protocol as we did with the synthetic images. Firstly, we performed an experiment on a subset of 500 images captured by Cameras Cl and Cr. This subset contained 197 positive images and 303 negative images. Our analysis yielded a precision score of 87.66% and a recall score of 89.8%. These results demonstrate that our model did not overfit to the synthetic labeled dataset during training, and it has the ability to generalize to real scenarios, effectively bridging the domain gap in this task. Furthermore, we conducted an additional experiment using all the real images captured by cameras A, B, Cl, and Cr (see Fig. 4). This dataset contained 1000 images, with 197 positive door images retained from the previous experiment. Our analysis revealed a precision score of 55.3% and a recall score of 89.8%. We observed a decrease in precision compared to the results presented in Table 10, which is expected given the wider negative set used in this experiment.

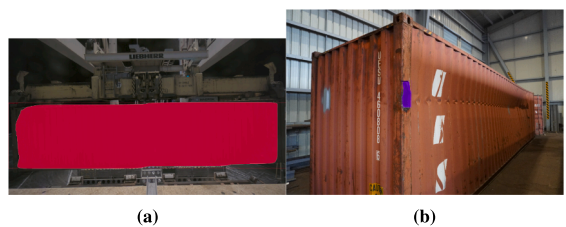

**Fig. 10.** Testing domain gap with real images: **(a)** Clear full container segmentation using a familiar setup, and **(b)** Limited segmentation of perforation damage with an unseen viewpoint.

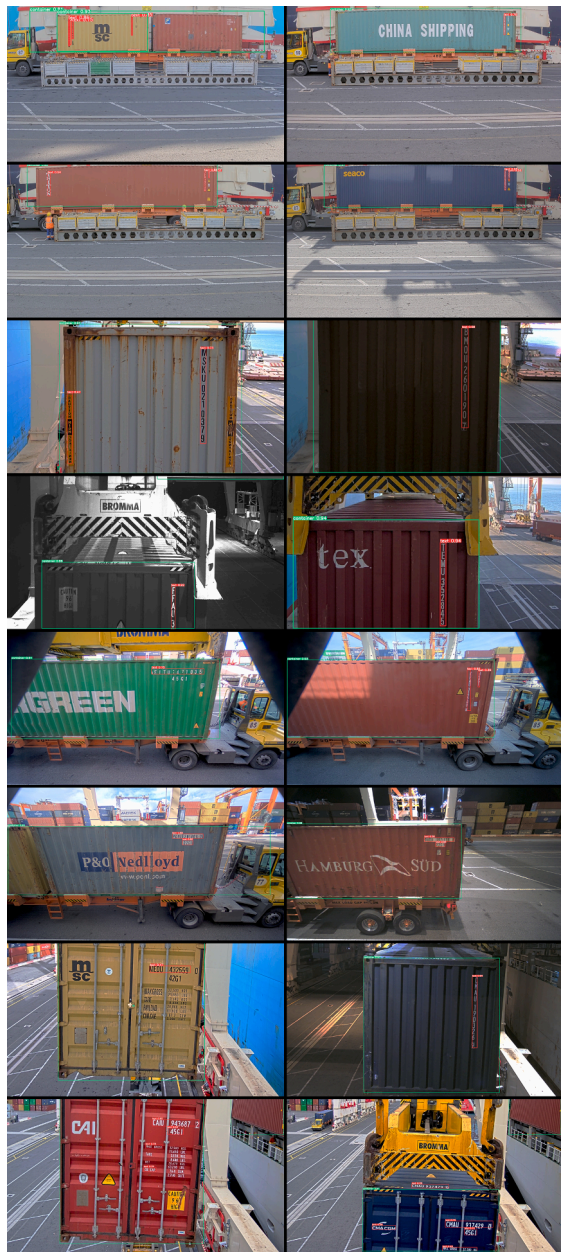

**Fig. 11.** Results of real images in order to test the domain gap for the detection of container and text classes marked with green and red colours respectively.

To sum up, our research offers an initial framework for addressing scenarios where public datasets are not available to researchers. It is crucial to note that the presented methodology should not be considered a substitute for real data, but rather as a supplementary approach to support areas lacking in such resources. Furthermore, it is important to recognize that the efficacy of our methodology may be impacted by variations in factors such as the task, point of view, or environment. To account for such potential variations, our methodology is designed to be adaptable and customizable. By demonstrating the effectiveness of our approach in most of the designated tasks, we have established its potential for application in other harbour contexts with unique requirements and tasks. To the tasks that has been evaluated empirically, we anticipate that it will be effective in this context as well. However, if a significant domain gap exists for a particular user's task, our methodology can still be employed to generate new types of tasks.

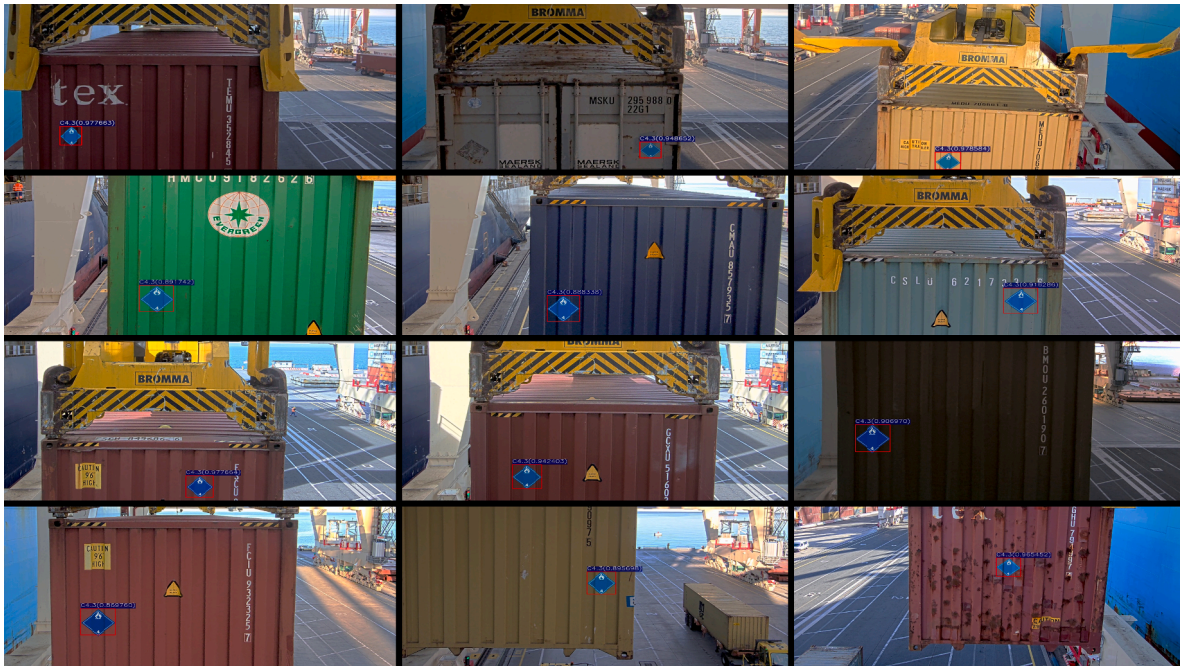

**Fig. 12.** Results of real images in order to test the domain gap for the detection of container and text classes marked with green and red colours respectively.

## 7. Conclusions

This study is the first step towards improving visual inspection operations in port facilities regardless of the limitations imposed by the acquisition of real labelled images. Therefore, this work presents a methodology to generate and build a synthetic labelled dataset for training visual inspection tasks for containers with enough generality to use with real data. Our set up replicates an STS scenario from Luka Koper port. However, our methodology allows researchers to generate any set ups regarding of their configurations and also allows to generate any situation of interests. Synthetic labelled datasets are a feasible alternative to train state-of-the-art neural models to being applied into real-world problems.

In addition, we provide the first free and open synthetic labelled dataset with multiple visual inspection task of containers for the research community with more than 10,000 images. Thus, not only providing a complete labelled dataset but a methodology to generate even more data regarding of the use case in this field that is so scarce of data. We also proposed multiple tasks which are found recurrently in the state of the art, and we are able to solve them using our methodology. For each task we provide balanced data and due to the randomness in the methodology we also provide a great data variability.

Conclusively, our work presents an advancement by introducing a methodology that enhances the precision of freight containers visual inspection task in the absence of publicly available datasets. Our dataset mirrors a real-life scenario, and our experiments demonstrate that training with synthetic data can yield favourable outcomes. It is important to emphasize that the methodology presented in this study should not be viewed as a replacement for real data but rather as a complementary strategy to supplement areas where such resources are scarce. Nevertheless, it serves as a promising starting point for further investigation and development. We aspire that our research motivates other researchers to investigate comparable strategies to improve container identification systems in other domains.

Future work includes the extension of the methodology by adding more visual tasks and improving the variety of the existing ones, as some tasks requires a more detailed definition. In addition, our future research endeavors will concentrate on tasks aimed at enhancing the detection and segmentation of damages in real tagged images, with the objective of conducting a further comprehensive evaluation by proposing a new dataset that could serve as a guideline. This will help improving the generalization capability of our dataset. We also plan to explore a semi-supervised training approach with synthetic images in order to improve the results using real unlabelled data. In addition, we will explore augmentation procedures in order to increment the dataset variability. Finally, we plan to redesign state-of-the-art architectures in order to reduce as much as possible the domain gap between real and synthetic images and introduce multi-tasks models that could be used to solve multiple proposed tasks at once in a single architecture.

## CRediT authorship contribution statement

**Guillem Delgado:** Conceptualization, Methodology, Software, Validation, Resources, Investigation, Data curation, Writing – original draft, Writing – review & editing, Visualization. **Andoni Cortés:** Conceptualization, Methodology, Software, Validation,

Investigation, Data curation, Writing – original draft, Writing – review & editing, Visualization. **Sara García:** Methodology, Software, Resources, Data curation, Writing – original draft, Visualization. **Estíbaliz Loyo:** Conceptualization, Writing – original draft, Project administration, Funding acquisition. **Maialen Berasategi:** Methodology, Software, Validation, Investigation, Data curation, Writing – original draft, Writing – review & editing. **Nerea Aranjuelo:** Conceptualization, Writing – original draft, Writing – review & editing.

**Declaration of Competing Interest**

The authors declare that they have no known competing financial interests or personal relationships that could have appeared to influence the work reported in this paper.

**Data availability**

SeaFront dataset is available in: https://datasets.vicomtech.org/di21-seafront/readme.txt.

**Acknowledgments**

This work has been partially done under the frame of the project 5GLOGINNOV (Grant agreement ID: 957400) funded by the European Commission under the H2020-ICT-2018-20 programme, within the topic ICT-42-2020 - 5G PPP – 5G core technologies innovation. Views and opinions expressed are however those of the authors only and do not necessarily reflect those of the European Union or CINEA. Neither the European Union nor the granting authority can be held responsible for them.